%% file: Formatting-Instructions-LaTeX-2022.tex
\def\year{2022}\relax
\documentclass[letterpaper]{article} 
\usepackage{aaai22}  
\usepackage{times}  
\usepackage{helvet}  
\usepackage{courier}  
\usepackage[hyphens]{url}  
\usepackage{graphicx} 
\usepackage{natbib}  
\usepackage{caption} 
\DeclareCaptionStyle{ruled}{labelfont=normalfont,labelsep=colon,strut=off} 
\usepackage{algorithm}
\usepackage{algorithmic}

\usepackage{amssymb}        

\usepackage[table,dvipsnames]{xcolor}
\usepackage{multirow}
\usepackage{xargs}
\usepackage{graphicx}
\usepackage{courier}
\usepackage{soul}
\newcommand*\mystrut[1]{\vrule width0pt height0pt depth#1\relax}

\usepackage{subcaption}
\usepackage[belowskip=5pt]{subcaption}

\usepackage{sidecap}

\usepackage[linewidth=1.2pt,linecolor=red]{mdframed}

\input{macros}

\usepackage{listings}
\floatstyle{ruled}
\newfloat{listing}{tb}{lst}{}
\floatname{listing}{Listing}
\usepackage{xcolor}
\usepackage{url}            
\usepackage{booktabs}       
\usepackage{amsfonts}       
\usepackage{nicefrac}       
\usepackage{microtype}      
\usepackage{lipsum}

\usepackage{amsmath}

\usepackage[english]{babel}

\usepackage{makecell}
\usepackage{multirow}

\usepackage{arydshln}

\title{Transcribing Natural Languages for the Deaf via Neural Editing Programs}
\author{
Dongxu Li$^{1,2,\dagger}$, Chenchen Xu$^{1,2}$, Liu Liu$^{3}$, Yiran Zhong$^{4,\dagger}$, \\Rong Wang$^{1}$, Lars Petersson$^{1,2}$, Hongdong Li
$^{1}$}
\affiliations{$^\dagger$corresponding authors: \{dongxu.li@anu.edu.au,~zhongyiran@sensetime.com\}\\\vspace{0.5em}
$^{1}$The Australian National University,  $^{2}$DATA61-CSIRO, $^{3}$Huawei Cyberverse Lab, $^{4}$SenseTime}
\usepackage{bibentry}

\begin{document}

\maketitle

\input{0-abstract}
\input{1-intro}
\input{2-relatedwork}
\input{3-methodology}
\input{4-experiments}
\input{5-conclusion}

\bibliography{egbib}

\end{document}

%% file: macros.tex
\newcommand{\ie}{\textit{i.e.}}
\newcommand{\eg}{\textit{e.g.}}

\definecolor{navyblue}{rgb}{0.17, 0.22, 0.71}
\definecolor{lava}{rgb}{0.81, 0.06, 0.13}
\definecolor{darkgreen}{HTML}{63A987}
\definecolor{codepink}{HTML}{e45da1}
\definecolor{codebrown}{HTML}{c7a763}


%% file: 0-abstract.tex
\begin{abstract}

%
%
%
This work studies the task of \emph{glossification}, of which the aim is to {\em transcribe} natural spoken language sentences for the Deaf (hard-of-hearing) community to ordered sign language glosses.
%
%
%
Previous sequence-to-sequence language models trained with paired sentence-gloss data often fail to capture the rich connections between the two distinct languages, leading to unsatisfactory transcriptions.
%
%
%
We observe that despite different grammars, glosses effectively simplify sentences for the ease of deaf communication, while sharing a large portion of vocabulary with sentences.
%
%
This has motivated us to implement glossification by executing a collection of editing actions, \eg~word addition, deletion and copying, called \emph{editing programs}, on their natural spoken language counterparts.
Specifically, we design a new neural agent that learns to synthesize and execute editing programs, conditioned on sentence contexts and partial editing results.
The agent is trained to imitate minimal editing programs, while exploring more widely the program space via policy gradients to optimize sequence-wise transcription quality. 
Results show that our approach outperforms previous glossification models by a large margin, improving the BLEU-4 score from 16.45 to 18.89 on RWTH-PHOENIX-WEATHER-2014T and from 18.38 to 21.30 on CSL-Daily.
Implementations will be made public.
\end{abstract}

%% file: 1-intro.tex
\section{Introduction}
Glossification is the task of transcribing natural language sentences into glosses, the written form of sign languages~\cite{johnston2007australian}. Each sign gloss is usually a word that relates to a sign gesture.
Glossification has important applications in automating deaf-hearing communication.
Such a transcription step is considered as a necessary precursor to translating natural languages into videos of sign gestures~\cite{stoll2020text2sign,korte2020plan}, thus alleviating the communication obstacles that the deaf and hard-of-hearing community members face and maximizing their performance in careers and other social engagement~\cite{yin2021including}.

Gloss sequences follow their own ordering rules, and usually consist of fewer tokens than their natural language counterparts. 
For instance, the English sentence \emph{``Do you like to watch baseball games?''} transcribes to American Sign Language (ASL) glosses \emph{``baseball watch you like?''}. 
Such discrepancy in grammar requires glossification models to jointly represent both language sources in the embedding space and properly build mappings in-between, yielding a challenging sequence learning problem.


Previous glossification approaches~\cite{stoll2020text2sign,zhangapproaching} take as input the natural language sentences and directly predict gloss sequences as output.
However, gloss annotations are expensive to obtain due to the required expertise in sign language.
As a result, the performances of models suffer from the limited amount of available parallel samples.
Considering this data-scarce situation, we aim to improve glossification quality by effectively exploiting the syntactic connections between sentences and glosses.

\begin{figure}
    \centering
    \includegraphics[width=1.03\linewidth]{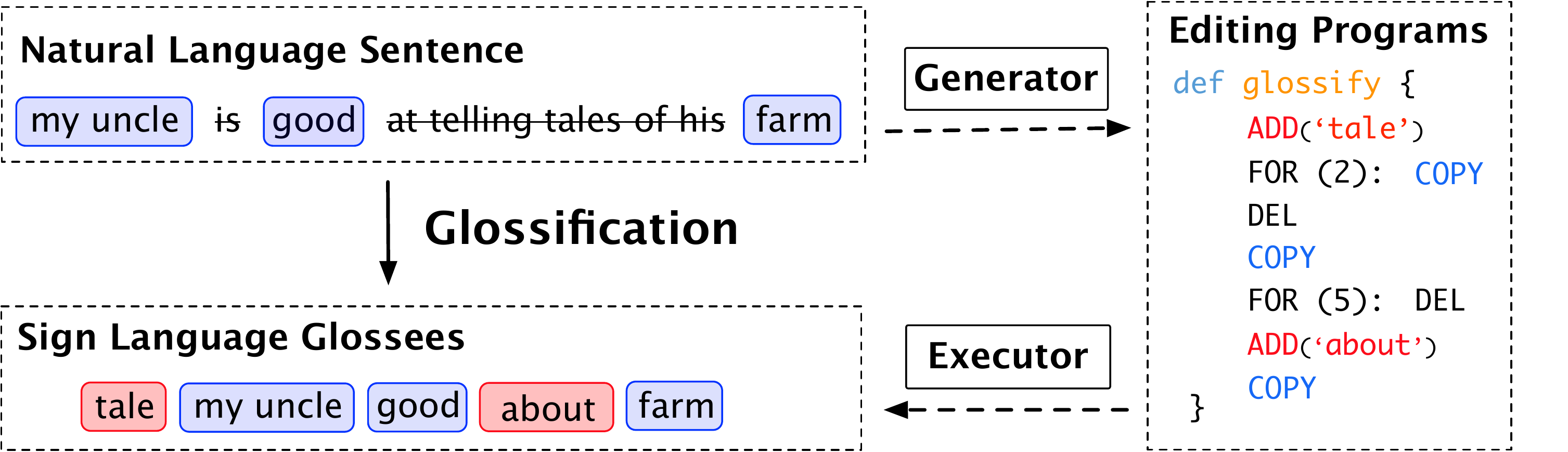}
    \caption{We study the problem of \emph{glossification}, which aims to transcribe natural language sentences into sign language glosses. In particular, we design a neural agent that generates and executes \emph{editing programs} on the natural language sentences to obtain glosses.}
    \vspace{-5mm}
    \label{fig:teaser}
\end{figure}


Specifically, we notice that despite grammar gaps, glosses largely share a common vocabulary with sentences.
In addition, glosses usually simplify sentences by keeping key content words and discarding those otherwise, for the ease of signing.
These two observations lead us to the following perspective: instead of predicting glosses directly, our model derive transcriptions by making changes to the original sentence.
%
%
In particular, we propose a new neural agent that learns to explicitly synthesize an ordered collection of editing actions for each input sentence, called an \emph{editing program}.
An editing program is composed of editing actions that remove, keep and add words based on the input sentence.
%
%
By sequentially executing editing actions on the input sentence, the agent obtains glosses as output.
%

Our proposed agent consists mainly of two components that collaborate with each other: a \emph{program generator} that at each time step predicts an editing action, and a \emph{program executor} that applies actions on the input sentence to obtain editing results, namely glosses.
To facilitate the communication between the two modules, we further introduce a new attention mechanism, called \emph{editing causal attention}, allowing the generator to attend to history partial glosses while preserving the auto-regressive property of the model.
%
%

Concretely, to learn the mapping from sentences to editing actions, the generator follows a typical transformer encoder-decoder structure, taking sentences as input. 
Different from previous approaches, instead of producing glosses directly, the generator synthesizes \emph{partial editing programs} at each step as output.
However, since the generator module receives editing labels as supervision, it learns mainly action labels yet not effectively leveraging the execution results, \ie~glosses, when making editing decisions.
To address this issue, in our design we enable the executor module to communicate the glosses output as feedback to the generator to guide further action predictions.
Particularly, at each step the executor first applies the predicted actions on the sentence to obtain partial gloss output, it then summarizes the glosses using an extra encoder.
%
%
Finally, before the generator predicts an action, it communicates with the executor regarding the current editing results via the proposed editing causal attention.
The editing causal attention is a variant of vanilla decoder attention, with the critical difference in that \emph{the number of masked tokens is determined dynamically by the editing history}.
With editing causal attention, we ensure the generator effectively attends to known partial editing results, thus better utilizing the program semantics for predicting further editing actions.

The agent is first trained to imitate minimal editing programs, obtained using a dynamic programming procedure similar to the Levenshtein distance algorithm~\cite{schutze2008introduction}.
However, we notice our agent gets overly-penalized due to the issue of program aliasing~\cite{bunel2018leveraging}: while multiple editing programs result in equivalent glosses yet all except the one provided as target are considered incorrect.
To alleviate this issue, we adopt a policy-gradient method~\cite{luo2020better} to reward the agent with semantically correct transcriptions, which we refer as \emph{peer-critic}.
The peer-critic method takes sequence-metrics as rewards, such as BLEU, and uses average rewards of peer samples as a baseline to reduce variance.
Combining the imitation and reinforcement learning strategies, our agent achieves significantly better glossification results than existing methods.
%

\noindent\textbf{Contributions.} Our main contribution are as follows. (i) We introduce a novel glossification glossification method via sequential executions of editing actions. Such a formulation effectively enables to exploit syntactic connections between sentences and glosses, making our method stand out from peers that rely on conventional machine translation pipeline; (ii) We design a causal editing attention module, a variant of typical transformer decoder attention where the number of masked tokens are determined dynamically based on the produced partial glosses.
In this way, we inform the generator of history execution results before making future decisions.
(iii) We optimize our agent by imitating minimal ground-truth editing programs, while also encouraging it to explore wider program space to counteract the effect of program aliasing.
(iv) Experiments on public datasets using sign languages from different regions show clearly preferable transcription quality from our system, both quantitatively and qualitatively through human evaluations with deaf involvement.

%
%
%
%
%

%

%
%
%



%% file: 2-relatedwork.tex
\section{Related Work}
\noindent\textbf{Sign Language Recognition, Translation and Production.} Most research works in automated sign language interpretation~\cite{yin2020better} aim at recognizing~\cite{li2020word,li2020transferring,albanie2020bsl,Min_2021_ICCV} and translating~\cite{cihan2018neural,NEURIPS2020_8c00dee2,2021Improving} visual sign gestures into sentences.
However, to facilitate two-way deaf-hearing communication, it is also necessary to translate spoken sentences to sign gestures in videos or animations for the deaf~\cite{korte2020plan}.
In this regard, the recent work~\cite{stoll2020text2sign} first translates sentences to glosses, which are later used to produce continuous sign language videos~\cite{saunders2021mixed}.
Our work follows this pipeline while for the first time, formalizing glossification as a standalone sign language interpretation task. 
Different from~\cite{stoll2020text2sign} that adapts a neural machine translation approach, we propose to use editing labels to bridge the gap between two linguistic sources, achieving superior glossification results.


\noindent\textbf{Neural Program Synthesis.}~~Program synthesis techniques aim to generate programs that satisfy given specifications~\cite{NEURIPS2020_cd0f74b5,pu2020program}, either in natural languages or as a set of example inputs and desired outputs.
The advantage of neural programs is their flexibility in modeling compositional structures in textual and visual data, thus are widely applied in various domains, including string manipulation~\cite{Reed:2016}, sentence simplification~\cite{dong2019editnts}, semantic parsing~\cite{shin2019program} and shape generation~\cite{NEURIPS2018_67880768,tian2018learning}. 
%
%
Inspired by these works, we use editing programs to exploit syntactic connections between sentences and glosses.
Different from previous models that directly predict glosses~\cite{stoll2020text2sign}, we instead obtain glosses as the result of executing editing programs on their sentence counterparts.
In this way, our model better utilizes relations between sentences and glosses by explicitly copying or removing words.
In addition, editing programs define each glossification step as an action, thereby, they are easier to interpret than results from black-box sequence-to-sequence machine translation models.
%
%
%

\input{tables/dsl}

\noindent\textbf{Learning Symbolic Operations on Text Sequences.}~There are also prior approaches relying on predicting symbolic operations for other sequence learning problems.
In particular, the dependency parsing models proposed in~\cite{chen2014fast,dyer2015transition}  predict transitions between initial and terminal configurations, which are then used to derive a target dependency parse tree.
The models in~\cite{alva2017learning,dong2019editnts} predict simplification operations to transform complex sentences into simple ones.
Inspired by these works, we design our program synthesis model that takes natural language sentences as input and learns to perform editing operations on them to obtain glosses.


%


%% file: tables/dsl.tex
\begin{table*}
\vspace{-0.5em}
  \caption{The syntax of domain specific language (DSL) used by editing programs in EBNF notation~\cite{visser1997syntax}. We represent non-terminal symbols on the left and production rules on the right.}
  \label{tbl:dsl}
  \centering
  \small
  {
  \begin{tabular}{rcc}
\toprule   
  Program & $\rightarrow$ & Statement; Program~$\mid$ $\varnothing$\\
  AtomicStatement & $\rightarrow$ & \texttt{ADD}(Token) $\mid$ \texttt{DEL}(PositionPointer) $\mid$ \texttt{COPY}(PositionPointer) $\mid$ \texttt{SKIP}\\
  Statement & $\rightarrow$ & \texttt{For}(RepeatParam); AtomicStatement; \texttt{EndFor} $\mid$ AtomicStatement\\
  Token & $\rightarrow$ & $t,\,t\in V$, where $V$ is the shared vocabulary of glosses and sentences.\\
  PositionPointer & $\rightarrow$ & $i,\,i\in\mathbb{N}_{\ge 0}$\\
  RepeatParam & $\rightarrow$ & $r,\,r\in\mathbb{N}^+$\\
   \bottomrule
  \end{tabular}
  }
\end{table*}

%% file: 3-methodology.tex
\section{Methodology}
In this section, we present the main technical contributions of our proposed approach.
First, we describe the \textbf{\emph{definition of editing programs}} and also the way we construct ground-truth editing programs for training.
Then, we detail the proposed \textbf{\emph{architecture of the generator and executor}} modules.
We also explain how these two modules communicate with each other, via a novel \textbf{\emph{ editing causal attention}} mechanism.
Finally, we introduce the \textbf{\emph{imitation and reinforcement learning strategies}} we adopt to train the glossification agent and alleviate the issue of program aliasing.

\subsection{Editing Programs for Glossification}\label{sec:edit-prog}


\noindent\textbf{Problem definition.}
Given $\boldsymbol{x} = [x_1,...,x_m]\in\mathcal{X}$ a natural language sentence with $m$ words from a vocabulary $V$, and $\boldsymbol{y} = [y_1,...,y_n]\in\mathcal{Y}$ the transcription with $n$ glosses from the same shared vocabulary $V$, an {\em editing program synthesis} approach aims to compute an valid editing program $\boldsymbol{z}\in\mathcal{Z}:\mathcal{X}\rightarrow\mathcal{Y}$ which transforms $\boldsymbol{x}$ to $\boldsymbol{y}$, \ie, $\boldsymbol{z}(\boldsymbol{x})=\boldsymbol{y}$. Note that $\boldsymbol{z}$ is not unique and there may exist multiple programs satisfying the input-output specification. 




\noindent\textbf{Definition of editing programs.}
The syntax of the domain specific language (DSL) for editing programs is given in Table~\ref{tbl:dsl}.
Specifically, each program $\boldsymbol{z}$ contains a variable number of program statements, including four atomic statements (or \emph{editing actions}) and a looping construct.

In terms of the atomic statements, we define (i) \texttt{ADD}(\texttt{w}), which selects a token \texttt{w} from the vocabulary and appends \texttt{w} to the gloss sequence $\boldsymbol{y}$. The sentence $\boldsymbol{x}$ remains intact when an \texttt{ADD} action is applied; (ii) \texttt{DEL}($k$), which removes the word $x_k$ from the sentence; (iii) \texttt{COPY}($k$), which keeps the word $x_k$ from the sentence and appends it to $\boldsymbol{y}$, for example; (iv) \texttt{SKIP}, which discards remaining sentence tokens and completes the glossification procedure.

We also introduce a looping construct, \texttt{For}($r$), that applies an atomic statement for $r$ repetitions. Benefits for including the \texttt{For} statement in the editing program are threefold. First, it captures the regularity when several consecutive words in the sentence are handled by the same atomic statement. Second, it reduces the length of programs and eases the difficulty during long-range inference. Third, since gloss sequences are usually shorter than their sentence counterparts, the number of \texttt{DEL} actions to apply is larger than other actions. In this regard, the \texttt{For} statement alleviates the challenge of synthesizing programs with imbalanced action classes.\footnote{In the rest of the manuscript, we omit the action parameters when it is unambiguous from the context.}
\noindent\textbf{Minimal editing program.}
As aforementioned, our agent learns to predict editing actions to derive glosses.
To achieve this, we provide expert editing programs to demonstrate program syntax and semantics to the agent.
In this regard, we first design a dynamic programming algorithm to compute \emph{minimal editing programs} for each sentence-gloss pair.

Given a sentence-gloss pair, a minimal editing program is the one that consists of the least number of \texttt{ADD} and \texttt{DEL} actions to transform a sentence to its gloss transcription.
Specifically, we adapt the procedure to compute Levenshtein distances~\cite{schutze2008introduction} while discarding the substitution actions, thereby avoiding the quadratic growth of the number of editing actions with the vocabulary size.
We first compute the minimal editing distance~\cite{schutze2008introduction} between the sentence $\boldsymbol{x}$ and the glosses $\boldsymbol{y}$, and then extract actions from the trajectory with the minimal editing distance. 
When there existing multiple trajectories of the same number of editing actions, we priortize \texttt{ADD} over \texttt{DEL} to ensure the uniqueness of the minimal editing program. 
Finally, we compress the identical consecutive actions by the \texttt{For} statement.
\begin{figure*}
    \centering
    \includegraphics[width=0.9\linewidth]{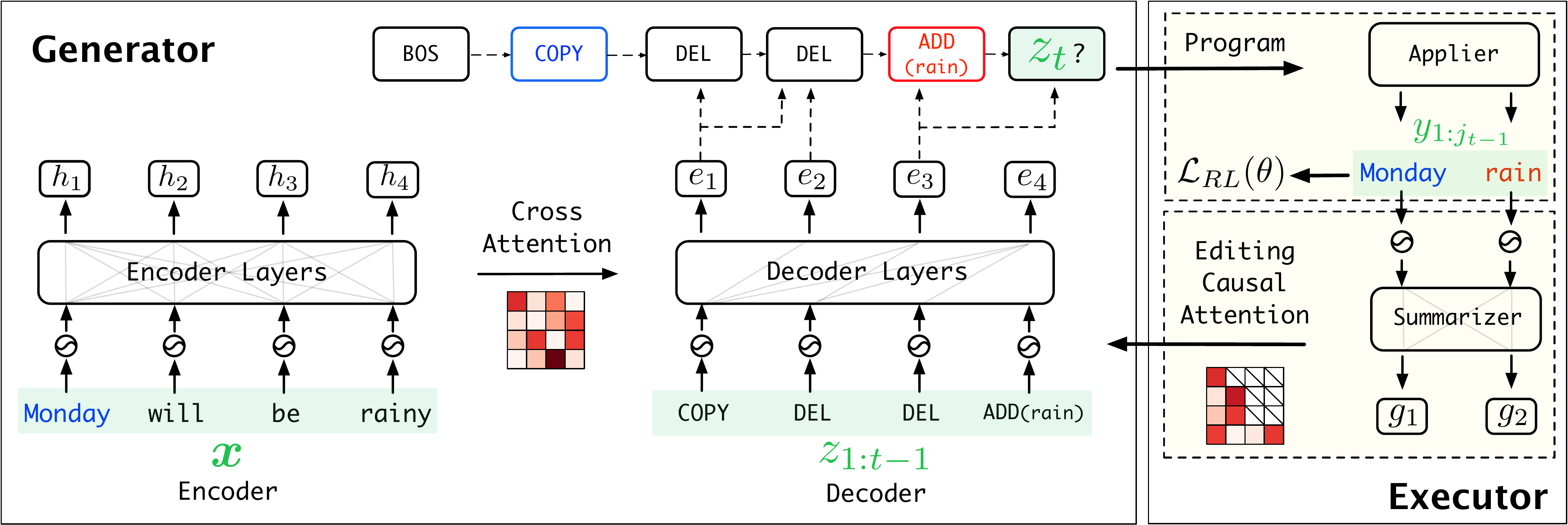}
    \caption{Our glossification model consists of two main modules: a \emph{generator} and an \emph{executor}.
    The generator predicts editing actions from the input sentence.
    The executor applies the program to derive glosses and provides execution feedback, which is then communicated with the generator via two channels: the editing causal attention mechanism and the peer-critic objective.}
    \label{fig:architecture}
\end{figure*}

\subsection{Neural Editing Program Synthesis and Execution}\label{sec:model}

%

An overview of our glossification model is shown in Fig.~\ref{fig:architecture}.
Given a sentence in natural language $\boldsymbol{x}=[x_1,...,x_m]$, our model predicts an editing program $\boldsymbol{z}$ to glossify $\boldsymbol{x}$ to $\boldsymbol{y} = [y_1,...,y_n]$ by modeling the conditional distribution $P(\boldsymbol{z}|\boldsymbol{x})$,
\begin{equation}~\label{eq:prob}
P(\boldsymbol{z}|\boldsymbol{x})=\prod_{t=1}^{|\boldsymbol{z}|}P(z_t|~\boldsymbol{x},\,y_{1:j_{t-1}},\,z_{1:t-1}).
\end{equation}
Particularly, at each time step $t$, we use a {\em generator} to predict the next statement $z_t$ considering (i) the natural language sentence $\boldsymbol{x}$; (ii) history gloss outputs $y_{1:j_{t-1}}$, where $j_{t-1}$ denotes the length of output glosses until time $t-1$; and (iii) history editing statements $z_{1:t-1}$ until time $t-1$. 
To effectively utilize history editing results, we also design an {\em executor} taking $\boldsymbol{x}$ and partial statements $z_{1:t-1}$ as input, then derives and summarizes history glosses $y_{1:j_{t-1}}$.

Since our editing program generation procedure is partially conditioned on the history execution results, we first introduce the program executor followed by the generator and their communication mechanism in-between.
%
%
%

\subsubsection{Program Executor.}
Given a natural language sentence $\boldsymbol{x}$ and a synthesized (partial) program $z_{1:t-1}$, our program executor first applies the editing statements on the sentence $\boldsymbol{x}$ to obtain the gloss output $y_{1:j_{t-1}}$.
To achieve this, we maintain an \emph{executor pointer} $k$ that holds the index of the current word to edit in $\boldsymbol{x}$.
Before the agent applies a \texttt{COPY} or \texttt{DEL} action, it determines the word to edit based on the pointer value.
Rules to update $k$ are as follows. The pointer $k$ starts from $x_1$, the first element of $\boldsymbol{x}$. Each time a \texttt{DEL} or \texttt{COPY} action is executed, $k$ moves to the next position of $\boldsymbol{x}$ and points to $x_{k+1}$, indicating $x_k$ is either kept in the glosses $\boldsymbol{y}$ or discarded during the execution.
When an \texttt{ADD} action is applied, $k$ remains unchanged since no editing happens in the sentence $\boldsymbol{x}$. On encountering a \texttt{For} statement, $k$ moves forward by $r$ positions and points to $x_{k+r}$.
Glosses $y_{1:j_{t-1}}$ are then obtained by executing the partial program $z_{1:{t-1}}$ sequentially on the sentence $x$.


After obtaining the partial glosses, the executor summarizes the output $y_{1:j_{t-1}}$ and prepares for communicating this execution result with the generator for future predictions.
To achieve this, the executor represents the glosses in the embedding space.
%
%
%
%
%
In particular, we feed glosses $y_{1:j_{t-1}}$ to a number of Transformer encoder layers~\cite{vaswani2017attention} to obtain their hidden embeddings $g_{1:j_{t-1}}$. For the $l$-th encoder layer: 
\begin{equation}\label{eq:enc}\resizebox{1\hsize}{!}{${g}_1^{(l+1)},..., {g}_{j_{t-1}}^{(l+1)}=    \begin{cases}
      E_{y_1} + P_1,..., E_{y_{j_{t-1}}} + P_{j_{t-1}},&l=1,\\
      \textnormal{EncoderLayer}_{l}({g}_1^{(l)},...,{g}_{j_{t-1}}^{(l)}),&l>1;
      \end{cases}$}
\end{equation}
where $E\in\mathbb{R}^{|V|\times d_{\textnormal{model}}}$ and $P\in\mathbb{R}^{L_{\textnormal{max}}\times d_{\textnormal{model}}}$ are look-up tables that map the $i$-th gloss $y_i$ to its token embedding and sinusoidal positional encoding~\cite{vaswani2017attention}, respectively, with $L_{\textnormal{max}}$ the maximal input lengths and $d_{\textnormal{model}}$ the hidden dimension.
%
The EncoderLayer($\cdot$) is composed of self-attention layers and position-wise feed-forward networks to capture pairwise dependencies among feature embeddings.

\subsubsection{Program Generator.} We formulate the program generation procedure as a sequential prediction problem and employ an encoder-decoder model for generating programs.
%
%
In particular, the encoder of the generator takes as input the natural language sentence $\boldsymbol{x}$, and represents each word ${x}_i$ in the embedding space as ${h}_i\in\mathbb{R}^{d_{\textnormal{model}}}$, similar to the summarization procedure in Eq.~(\ref{eq:enc}).
The decoder models the conditional probability $P(\boldsymbol{z}|\boldsymbol{x})$ as in Eq.~(\ref{eq:prob}), while at the same time communicating with the executor regarding the history gloss output using an editing causal attention mechanism.

Specifically, at time step $t$, given the partial program $z_{1:t-1}$ and the hidden sentence representations $\boldsymbol{h}=[h_1,h_2,...,h_{m}]$ with $m$ the input length, the decoder first computes the representation of each statement in the editing history ${e}_i\in\mathbb{R}^{d_{\textnormal{model}}}$ using Transformer decoder layers~\cite{vaswani2017attention}.
\begin{equation}\label{eq:dec}\resizebox{0.99\hsize}{!}{$
    {e}_1^{(l^{\prime}+1)},...,{e}_{t-1}^{(l^{\prime}+1)}=    \begin{cases}
      E_{{z}_1} + P_1,..., E_{{z}_{t-1}} + P_{t-1},&l^{\prime}=1,\\
      \textnormal{DecoderLayer}_{l^{\prime}}({e}_1^{(l^{\prime})},...,{e}_{t-1}^{(l^{\prime})},\\~~~~~~~~~~~~~~~~~~~~~~~~~~~~~~{h}_1,...,{h}_{m}),&l^{\prime}>1;
      \end{cases}$}
\end{equation}
where $l^{\prime}$ is the index of decoder layers. The token embedding and positional encoding are similar to those in the encoder except that they apply to the editing history.
In addition to the two sub-layers as in the encoder, the decoder consists of an extra sub-layer, which performs
attention over the sentence representation $\boldsymbol{h}$ from the encoder.
In this way, the decoder considers the full context of the sentence input as well as the past editing statements when making further predictions.

\noindent\textbf{Editing causal attention for execution-guided generation.}
As described above, the program generator models the relation between the program $\boldsymbol{z}$ and the sentence $\boldsymbol{x}$.
However, there exists a clear gap in such a design that the partial gloss output $y_{1:t-1}$ is not fully utilized.
In other words, the generator largely ignores the semantics of the editing programs.
This is less desirable as the partial execution states provide useful guidance for predicting future program statements~\cite{chen2018executionguided}.
In particular for the glossification task, partial glosses provide helpful contexts for predicting the next editing action to take.
In light of this observation,  we develop an  \emph{editing causal attention}, a mechanism that effectively allows the generator to take into account history glosses in the followup program generation process.

The editing causal attention is a variant of masked multi-head attention~\cite{vaswani2017attention}.
Yet differently, since the model produces one sign gloss only when an \texttt{ADD} or \texttt{COPY} operation is applied, while the length the gloss output remains unchanged for \texttt{DEL} actions, the number of masks is dynamically determined based on the predicted editing history $y_{1:j_{t-1}}$.
In this way, we effectively prevent editing actions from peeking future glosses and preserve the auto-regressive property of the generator, thereby, preserving the auto-regressive property of the generator model.

Specifically, we maintain a \emph{generator pointer}, which points at the gloss sequence $\boldsymbol{y}$ and records $j_{t-1}$, the length of the current gloss sequence.
The pointer is initialized as zero, indicating that the history gloss is empty.
When the executor applies an \texttt{ADD} or \texttt{COPY} action, the generator pointer moves forward by one position, suggesting that one more gloss is produced.
On encountering a \texttt{For} statement, the pointer moves forward by the number of repetitions.
When a \texttt{DEL} action is applied, the pointer remains unchanged since no new gloss is produced.
During the teacher-forcing training, when predicting the editing action $z_t$ at time step $t$, gloss positions larger than $j_{t-1}$ are masked.
%
%

We add an editing causal attention layer on top of the last decoder layer. 
Specifically, the communication between the generator and the executor is achieved via a masked scaled dot-product attention $\mathcal{G}_{\mathrm{attn}}(\mathbf{Q},\mathbf{K},\mathbf{V})=\mathrm{softmax}(\mathbf{Q}\mathbf{K}^T/\sqrt{d})\mathbf{V}$, where $\mathbf{Q}$ represents transformed features of $e_{1:t-1}$ via a feed-forward layer and $\mathbf{K}$, $\mathbf{V}$ are those of $g_{1:j_{t-1}}$, $d$ is  the feature dimension of vectors in $\mathbf{K}$.
The attention matrix from the $\mathrm{softmax}(\cdot)$ function is masked such that illegal positions of future glosses are filled with $-\infty$, with the number of masked tokens determined as explained beforehand.
Finally, we add a single-layer feed-forward network as the classifier to predict the program statement $z_t$.
A visual example of learned editing causal attention map is shown in Fig.~\ref{fig:eca}.

\noindent\textbf{Handling looping statements.}~Generally, \texttt{For} statements in programming languages allow nested loops.
%
However, this creates additional complications in constructing deterministic minimal editing programs as supervision as well as during program generation.
To ease this complexity, we guide the generator with a restrictive usage of \texttt{For} constructs, allowing only atomic statements to repeat. During the program generation, this adaption effectively only requires the model to predict an atomic statement to apply, along with an integer denoting the number of repetitions.

%
%
%
%
%

\subsection{Learning to Imitate and Explore}\label{sec:optimize}
We provide the agent with initial task knowledge in an \emph{imitation learning} strategy, where the agent takes the teacher action $z_t^{*}$ at each time step to efficiently learn to imitate the minimal editing programs.
However, in practice, we notice the agent is overly-penalized due to the issue of program aliasing~\cite{bunel2018leveraging}.
Namely, multiple editing programs result in equivalent glosses yet all except the one provided as target are considered incorrect.
An obvious example as such in our case is that to \texttt{COPY} a token $x_k$ from the sentence $\boldsymbol{x}$ is equivalent to predict an \texttt{ADD} action with the same token $x_k$.

Inspired by recent works in visual captioning~\cite{rennie2017self,luo2020better}, we propose to combine the imitation learning (IL) objective with a policy gradient method (RL) that rewards the agent for correct gloss outputs.
In this way, we encourage the agent to not only exploit the expert knowledge in minimal editing programs, but also to explore more widely the program space for semantically-equivalent programs. Specifically, we compute the reward by evaluating the BLEU-4 score of the generated glosses with respect to the corresponding ground-truth glosses. The goal of RL objective is to minimize the negative expected reward.
The overall optimization objective combines IL and RL objectives as follows:
\begin{equation}\label{eq:rl_loss}
    \mathcal{L}(\theta)= \underbrace{\lambda\sum_{t=1}^T-z_t^*\mathrm{log}(q_t)}_{\mathcal{L}_{IL}(\theta)}\underbrace{\mystrut{2.85ex}-\mathbb{E}_{\Tilde{\boldsymbol{z}}\sim P_\theta(\boldsymbol{z}|\boldsymbol{x})}[r(\Tilde{\boldsymbol{z}})]}_{\mathcal{L}_{RL}(\theta)},
\end{equation}
with $q_t$ the output probability of ground-truth editing label $z_t^{*}$, $\Tilde{\boldsymbol{z}}=[\Tilde{z}_1,...,\Tilde{z}_t]$ the editing statement sampled from the model at time step $t$, $P_\theta(\boldsymbol{z}|\boldsymbol{x})$ the glossification model (generator and executor) parameterized by $\theta$ and $r(\cdot)$ the reward function, respectively. 
The hyperparameter $\lambda$ balances between the imitation learning and reinforcement learning objectives.
Since the $\mathcal{L}_{RL}(\theta)$ term in Eq.~(\ref{eq:rl_loss}) is non-differentiable \emph{w.r.t.} $\theta$, we then use the REINFORCE algorithm~\cite{williams1992simple} to compute the gradient with Monte-Carlo sampling,
\begin{equation}
    \nabla_{\theta}\mathcal{L}_{RL}(\theta)=-r(\Tilde{\boldsymbol{z}})\nabla_{\theta}~\mathrm{log}~P_{\theta}(\Tilde{\boldsymbol{z}}|\boldsymbol{x}
    ).
\end{equation}
%
In practice, we approximate the expected rewards with the average of $K=5$ samples from $P_\theta(\boldsymbol{z}|\boldsymbol{x})$.
To reduce the variance of the estimation, we follow the method as suggested in~\cite{luo2020better}, and further compute the reward function relative to a baseline $b$, resulting the reward as an advantage function. For each sample, its baseline is the average reward of the remaining $K-1$ peer samples. 
In the sequel, we refer this RL objective as the \emph{peer-critic} objective for simplicity.
We refer interested readers to~\cite{luo2020better} for more details.
%


%% file: 4-experiments.tex
\section{Experiments}\label{sec:exp}
\subsection{Implementation Details and Experiment Setup}\label{sec:exp_setup}
\noindent\textbf{Implementation.}~We implement our model with the framework \textsc{Fairseq}~\cite{ott2019fairseq} in \textsc{PyTorch}~\cite{NEURIPS2019_9015}.
To represent texts in the feature space, we use pre-trained German and Chinese embeddings~\cite{joulin2016fasttext}.
The generator consists of three encoder layers and one decoder layer; the executor consists of a single encoder layer. 
We adopt ten parallel heads in all the multi-head attention modules to learn diverse patterns.
During the training, we warm up the agent with the imitation learning objective for $25$ epochs so that it learns the syntax and semantic rules of DSL efficiently.
Then we add the peer-critic objective to encourage the agent to explore more widely the program space with semantically correct statements.
We optimize our model using the Adam optimizer~\cite{kingma2014adam}, with an initial learning rate of  $10^{-4}$ and a weight decay of $10^{-4}$.
All the hyperparameters are selected using the validation partition.
We train our networks for 150 epochs, which is sufficient for all the models to converge, each taking around 30 hours on a single NVIDIA P100 GPU. Code will be made public.

\noindent\textbf{Datasets.}~We evaluate our glossification approach on two widely-used public datasets, including RWTH-PHOENIX-Weather 2014T (RPWT) dataset~\cite{cihan2018neural} and CSL-Daily~\cite{zhou2021improving}, the only two existing datasets that provide parallel sentence-gloss annotations for large-scale training and inference.

Specifically, \textbf{RPWT} has glosses in German Sign Language (GSL/DGS) and sentences in German, while \textbf{CSL-Daily} contains Chinese Sign Language (CSL) glosses paired with sentences in Chinese. On both datasets, we follow the public data partition protocol, with $7,096$, $519$, $642$ sentence-gloss pairs for training, validation and testing on RPWT and $18,401$, $1,077$ and $1,176$ pairs, respectively for CSL-Daily.
Using these two datasets, we validate and demonstrate the potentials of our approach to generalize to sign languages from different geographic regions.

%
%

\noindent\textbf{Metrics}. Our method achieves glossification by synthesizing editing programs.
To validate the proposed method, we evaluate two aspects of the synthesized editing programs.
\begin{itemize}
    \item \textbf{Program alignment.} The predicted program is a \emph{perfect alignment} if it is identical to the minimal editing program. When the alignment is not perfect, we compute the \emph{program error rate} (PER) to measure the alignment between the synthesized program and the minimal editing program, which adapts word error rate (WER)~\cite{hinton2012deep} used in speech recognition research and computes alignment between editing statements.
    \item \textbf{Generalization.}~The predicted program is a \emph{generalization} if it satisfies the input-output specification~\cite{chen2018executionguided}. For glossification, this requires comparing the predicted glosses with the ground-truth ones.
    In this regard, we use the BLEU~\cite{papineni2002bleu} and ROUGE-L~\cite{lin2004automatic} scores, two commonly adopted measurements for sequences. BLEU-$n$ measures the precision of the sequence up to $n$-gram. ROUGE-L measures the F1 score based on the longest common sub-sequences between predictions and ground-truth glosses.
\end{itemize}


\input{tables/rpwt_results}
\input{tables/example-out}
\input{tables/ablation_lambda}
\subsection{Method Evaluation}
\label{sec:eval}
The aim of the evaluation is three-fold. 
First, we demonstrate the advantage of the proposed glossification approach via synthesizing editing programs.
This is achieved by comparing our method (with ablations) with previous glossification and related sequence learning methods quantitatively.
Second, we take RPWT as an example and analyze effects of important components and design choices.
%
%
Third, we validate that our approach improves the transcription quality for the deaf community.
We achieve this by human evaluations on the CSL-Daily dataset with deaf involvement.

\noindent\textbf{Competing Methods.}~We compare our approaches with two groups of competing methods.
(i) Previous glossification approaches, including Text2Sign~\cite{stoll2020text2sign} and Zhang et al~\cite{zhangapproaching}. Both works adopt a conventional encoder-decoder architecture, taking sentences as input and glosses as output.
In particular, Text2Sign uses GRU~\cite{chung2014empirical} as the underlying model and Zhang et al. use Transformers~\cite{vaswani2017attention}.
(ii) Related methods from other sequence learning tasks.
Considering the overlap between sentences and glosses, we compare with CopyNet~\cite{gu2016incorporating}, a general sequence learning method that selectively replicates segments from the input to the output. Such copying mechanism proves desirable for the glossification task.
Observing that glosses are usually shorter than their sentence counterpart, we also compare with a text simplification model EditNTS~\cite{dong2019editnts} that simplifies complex sentences by explicit editing operations.


\noindent\textbf{Quantitative comparison.}~Results are shown in Table~\ref{tab:rpwt}. The row of \emph{baseline} stands for the model without either the editing causal attention or the peer-critic objective.
As indicated in the table, our approach consistently outperforms previous glossification approach by a large margin.
Compared with approaches that directly predict glosses (Text2Sign, Zhang et al and CopyNet), our approach exploits the linguistic relation between sentences and glosses.
In this way, our agent effectively reuses content words from the sentence, thus obtaining superior transcription results.
Compared with EditNTS, our model follows a Transformer architecture and is equipped with the  novel editing causal attention module. 
The editing attention allows the generator to take into account of partial glosses as the feedback of program executions more flexibly than the hard attention in EditNTS.
A visual example of the editing causal attention is shown in Fig.~\ref{fig:eca}.
In addition, we observe that the peer-critic objective adversely impacts PER yet results in better sequence-level metrics.
This indicates that beyond following the minimal editing programs, our agent also searches for semantically correct editing programs.
These validate our motivations for the peer-critic objective.

\input{figs/attn}

%

\noindent\textbf{Qualitative results.}~Table~\ref{tbl:example-out} shows an example transcription produced by our model on RPWT. Our model succeeds in providing high-quality gloss sequences that match the ground-truth. Note that the example well demonstrates the issue of program aliasing: the minimal editing program consists of a \texttt{COPY} operation to obtain the word \emph{wechselhaft}, while our model decides to apply an \texttt{ADD} action alternatively. To obtain the correct glosses, our model then learns to apply one more deletion operation than the minimal editing program, in order to remove \emph{wechselhaft} from the original sentence.
%

\noindent\textbf{Model analysis and discussions.} (i) Without introducing \texttt{For} statements, the best BLEU-4 score drops to 18.13. This is because the number of \texttt{DEL} actions in the minimal editing programs increases, worsening the action imbalance issue.
(ii) We report the experiment results using different $\lambda$ values between multi-tasking objectives in Table~\ref{tab:abl_lambda}. It shows that introducing the peer-critic objective helps to improve the sequence metrics while does not guarantee fewer errors in modeling ground-truth programs.
(iii) We also experiment with using ROUGE-L scores and a combination of ROUGE-L and BLEU-4 as the reward function for the peer-critic objective. The best model achieves 50.17 in ROUGE-L score and 18.21 in BLEU-4, validating the effect of peer-critic. We report results using the BLEU scores as the objective because it provides more visible improvement on different metrics.

\noindent\textbf{Human evaluation with CSL users.} With the help of two judges, we report human evaluation results in Table~\ref{tab:human}. Both judges are from the deaf community and are native CSL users. During the evaluation, we ask the judges to rank the models based on (i) correctness: whether glosses follow correct grammars? This is necessary as regional sign languages dialect may lead to correct yet different transcriptions from the ground-truth; (ii) adequacy: how much intent from the original sentences is preserved? The average rankings from the two judges show that our system is overall preferred. However, transcription becomes harder when sentence lengths grow. As a result, the difference between models become relatively less evident. It is our future work to improve glossification performance especially on long sentence inputs.

%


%% file: tables/rpwt_results.tex
\begin{table}[t]
    \caption{Results of quantitative comparisons. We show metrics ROUGE-L (R-L), BLEU-3 (B-3) and BLEU-4 (B-4). We report official results or results from officially released models when possible, and use ($\dag$) to denote our reproduced results otherwise. Edit-Att abbreviates for the system with editing causal attention mechanism.
    }
    \centering
    {
    \begin{subtable}{0.48\textwidth}\centering{\begin{tabular}{lcccccccc}
    \toprule
    \textbf{Methods} & \textbf{PER}~$\downarrow$ & \textbf{R-L}~$\uparrow$ & \textbf{B-3}~$\uparrow$ & \textbf{B-4}~$\uparrow$ \\
    \midrule
    \multicolumn{5}{l}{
    \textbf{Previous Glossification Models}} \\
    Text2Sign & - & 48.10 & 21.54 & 15.26\\
      Zhang et al & - & 49.19 & 23.03 & 16.45\\
    \midrule
    \multicolumn{5}{l}{
    \textbf{Related Sequence Learning Models}} \\
    EditNTS
    &  - & 46.62 & 20.23 & 14.75 \\
    CopyNet$^{\dag}$ & - & 48.41 & 21.74 & 15.86\\
    \midrule
      Baseline  & 56.51 & 47.07 & 22.44 & 16.01 \\ 
      Edit-Att  & \textbf{55.24} & 49.66 & 24.93 & 18.07 \\
      Edit-Att + $\mathcal{L}_{RL}$ &  55.56 & \textbf{49.91} & \textbf{25.51} & \textbf{18.89} \\
    \bottomrule
    \end{tabular}}\caption{RWTH-PHOENIX-WEATHER-2014T (RPWT)}\end{subtable}
    }
      \begin{subtable}{0.48\textwidth}\centering
      \begin{tabular}{lcccccccc}
    \toprule
    \textbf{Methods} & \textbf{PER}~$\downarrow$ & \textbf{R-L}~$\uparrow$ & \textbf{B-3}~$\uparrow$ & \textbf{B-4}~$\uparrow$ \\
    \midrule
    \multicolumn{5}{l}{
    \textbf{Previous Glossification Models}} \\
    Text2Sign$^{\dag}$ & - & 49.19 & 25.46 & 18.80 \\
      Zhang et al$^{\dag}$ & - & 46.61 & 26.09 & 18.38 \\
    \midrule
    \multicolumn{5}{l}{
    \textbf{Related Sequence Learning Models}} \\
    EditNTS
    &  - & 51.89 & 28.21 & 19.88 \\
    CopyNet$^{\dag}$ & - & \textbf{52.84} & 28.77 & 20.24\\
    \midrule
      Baseline  & 48.29 & 50.31 & 25.91 & 17.97 \\ 
      Edit-Att  & \textbf{47.16} & 52.31 & 28.82 & 20.56 \\
      Edit-Att + $\mathcal{L}_{RL}$ &  48.30 & 52.78 & \textbf{29.70} & \textbf{21.30} \\
    \bottomrule
    \end{tabular}\caption{CSL-Daily}
      \end{subtable}\vspace{-4mm}
    \label{tab:rpwt}
\end{table}

%% file: tables/example-out.tex
\begin{table*}
\vspace{-0.5em}
  \caption{Example outputs of our glossification approach on the testing set of RPWT dataset.
  The \emph{prediction} rows show the synthesized programs and execution results.
  The \emph{reference} rows show minimal editing programs and ground-truth glosses. 
  We highlight tokens to \texttt{COPY} in  \textcolor{blue}{blue}, to \texttt{ADD} in  \textcolor{red}{red} and tokens to delete as to \st{strikeout}. We show correctly glossified 1-grams in \textcolor{ForestGreen}{green}.
  We add an integer $n$ after an action as an abbreviation for \texttt{For} statements with $n$ repetitions.
  }
  \label{tbl:example-out}
  \centering
  \small
  \resizebox{0.78\linewidth}{!}
  {
  \begin{tabular}{ll}
\toprule   
  Sentence & \makecell[l]{\textcolor{blue}{montag} \st{und} \textcolor{blue}{dienstag} \st{wechselhaft hier und da zeigt} \st{sich aber} \textcolor{blue}{auch die sonne} .\\ (monday and tuesday changeable here and there but the sun also shows up .)} \\
  \cmidrule{1-2}
  \multirow{2}{*}{Prediction} & {\textcolor{blue}{\texttt{COPY}} \texttt{DEL} \textcolor{blue}{\texttt{COPY}}} \textcolor{red}{\texttt{ADD}(wechselhaft)}
  \textcolor{red}{\texttt{ADD}(mal)} \texttt{DEL5} {\texttt{DEL2}} {\textcolor{blue}{\texttt{COPY}}} {\textcolor{blue}{\texttt{COPY}}} {\textcolor{blue}{\texttt{COPY}}} {\texttt{SKIP}} \\
  & \textcolor{blue}{montag dienstag} \textcolor{red}{wechselhaft mal} \textcolor{blue}{auch die sonne} 
 \\\cmidrule{1-2}
    \multirow{2}{*}{Reference} & \texttt{COPY} \texttt{DEL} \texttt{COPY}
    \texttt{COPY} \texttt{ADD}(mal) \texttt{DEL5} \texttt{DEL} \texttt{COPY} \texttt{DEL} \texttt{COPY} \texttt{SKIP} \\
  & \textcolor{ForestGreen}{montag dienstag wechselhaft mal auch sonne}
  \\
  \bottomrule
  \end{tabular}
  }
\end{table*}

%% file: tables/ablation_lambda.tex
\begin{table}
    \caption{Results with different $\lambda$ values (left) and with different metrics as rewards (right) on the RPWT dataset.
    }
    \resizebox{\columnwidth}{!}{%
    \begin{tabular}{ lr }   
\begin{tabular}{cccc} 
\toprule
\textbf{$\lambda$} &  \textbf{PER}~$\downarrow$ & \textbf{R-L}~$\uparrow$ &
 \textbf{B-4}~$\uparrow$ \\ \midrule
        {0.1} & 54.84 & 49.63 & 18.33  \\ 
        {0.5} & \textbf{55.56} & \textbf{49.91} & \textbf{18.89 }\\ 
        {1.0} & 55.68 & 50.46 & 18.60 \\
\bottomrule
\end{tabular} &  
\begin{tabular}{cccc} 
\toprule \textbf{Reward} &  \textbf{PER}~$\downarrow$ & \textbf{R-L}~$\uparrow$ &
 \textbf{B-4}~$\uparrow$ \\ \midrule
        {R-L} & 55.48 & \textbf{50.17} & 18.21\\ 
        {B-4} & 55.56 & 49.91 & \textbf{18.89} \\ {R-L+B-4} & \textbf{55.18} & 49.77 & 18.17 \\
\bottomrule
\end{tabular}\\
\end{tabular}
    \label{tab:abl_lambda}}
\end{table}
\begin{table}
    \caption{Average ranking for Correctness, Adequacy by two deaf volunteers on the CSL-Daily dataset on our methods, CopyNet (CN) and Zhang et al (ZH). We select ten short ($<$8 characters), medium (8$\sim$15 characters), long sentences each ($>$15 characters). BLEU-4 is also shown for the samples.
    }
    \centering
    \resizebox{\columnwidth}{!}{
    \begin{tabular}{cccccccccc} 
\toprule &  \multicolumn{3}{c}{\textbf{Correctness Rank~$\downarrow$}} & \multicolumn{3}{c}{\textbf{Adequacy Rank~$\downarrow$}} & \multicolumn{3}{c}{\textbf{BLEU-4}~$\uparrow$} \\ &
Ours & {CN} & {ZH} & {Ours} & {CN} & {ZH} & Ours & {CN} & {ZH}\\
\midrule
Short & \textbf{1.45} & 1.95 & 2.60 & \textbf{1.50} & 1.80 & 2.70 & \textbf{24.38} & 22.97 & 21.27 \\ 
Medium & \textbf{1.50} & 2.15 & 2.35 & \textbf{1.70} & 1.90 & 2.40  & \textbf{21.52} & 20.38 & 18.12 \\ 
Long & 1.95 & \textbf{1.90} & 2.15 & \textbf{1.85} & \textbf{1.85} & 2.30 & \textbf{19.77} & 19.42 & 17.87 \\
\bottomrule
\end{tabular}\label{tab:human}}
\end{table}

%% file: figs/attn.tex

\begin{SCfigure}[6][t]
\centering
\includegraphics[scale=0.24]{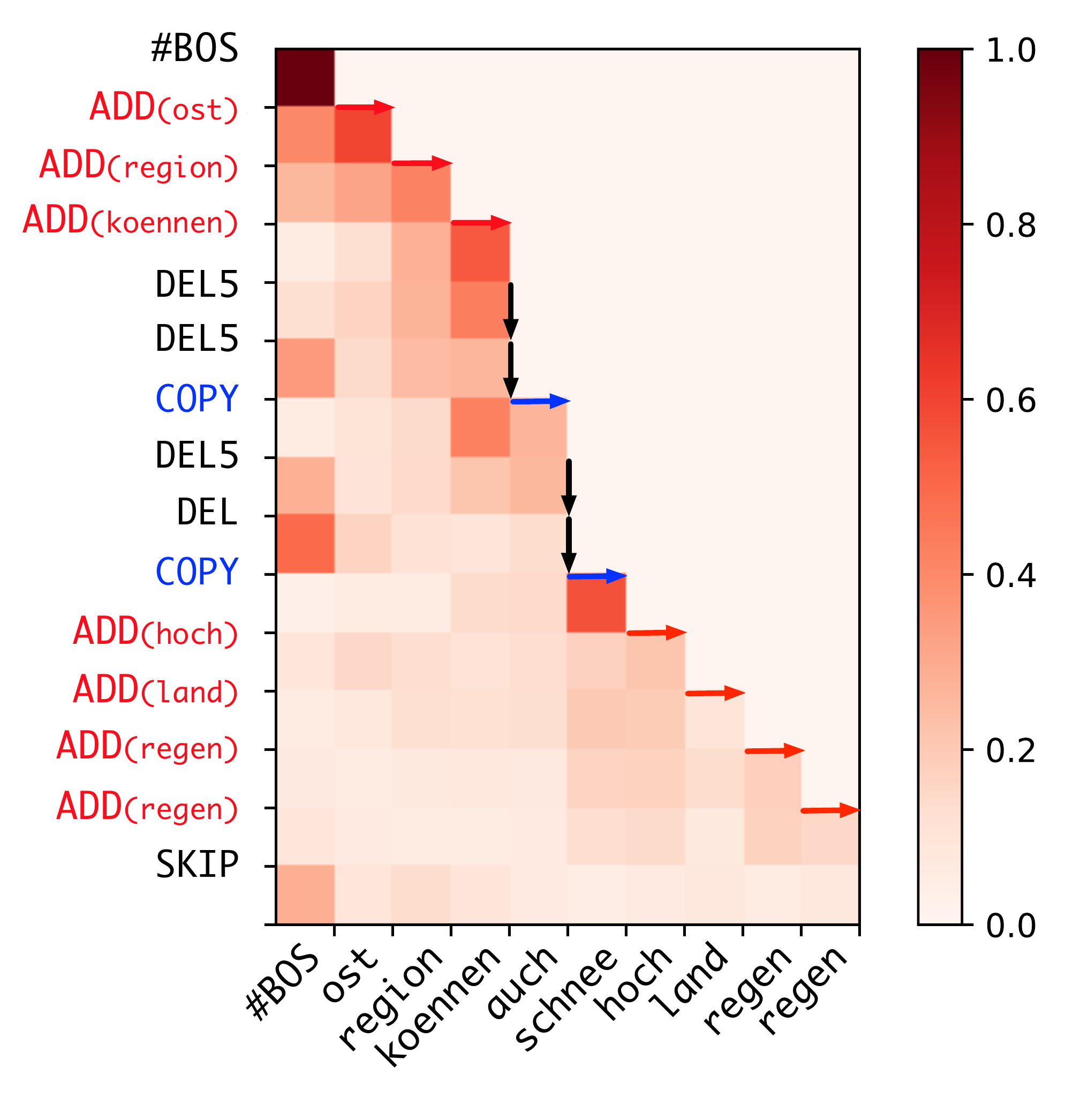}
\caption{A visual example of the editing causal attention. To prevent the generator from peeking at future glosses during training, we reveal the next gloss only when either a \texttt{COPY} or \texttt{ADD} action is executed. When \texttt{DEL} actions are executed, masks remain unchanged}
\label{fig:eca}
\end{SCfigure}

%% file: 5-conclusion.tex
\section{Conclusion}

In this work, we have proposed a new approach that transcribes natural language sentences into sign language glosses.
Instead of directly predicting glosses as output, our model learns to derive glosses as the result of editing the input natural language sentences.
This is achieved by a generator module which synthesizes the editing programs and an executor module that performs the actions.
These two modules communicate about the execution results via a new editing causal attention mechanism.
To account for the program aliasing issue, our agent learns to imitate the ground-truth action sequences while at the same time exploring the wider program space via a policy-based objective.
Our approach yields significantly better transcription quality on  commonly-adopted public sign language datasets, verified quantitatively and qualitatively by human evaluation with deaf involvement.
In addition, editing programs are more explainable than otherwise ``black-box'' type sequence-to-sequence models, offering a new perspective to the glossification task and visual sign language research in general.

%